\documentclass[conference]{IEEEtran}
\ifCLASSINFOpdf
\else
\fi
\usepackage[cmex10]{amsmath}
\usepackage{algorithm2e}
\usepackage[tight,footnotesize]{subfigure}
\usepackage{url}
\usepackage{subfig}
\usepackage{color,soul, graphicx}
\usepackage{kbordermatrix}

\usepackage{lipsum}

\newcommand\blfootnote[1]{%
  \begingroup
renewcommand*{\thefootnote}{\fnsymbol{footnote}}
\addtocounter{footnote}{-1}%
  \endgroup
}

\usepackage{amssymb}

\newtheorem{theorem}{Theorem}[section]

\newtheorem{definition}[theorem]{Definition}

\DeclareMathOperator*{\argmin}{\arg\!\min}

\IEEEoverridecommandlockouts

\begin{document}

\IEEEoverridecommandlockouts

\IEEEpubid{\makebox[\columnwidth]{978-1-4799-7560-0/15/\$31 \copyright 2015 IEEE \hfill} \hspace{\columnsep}\makebox[\columnwidth]{ }}

\title{Multivariate Time Series Classification Using Dynamic Time Warping Template Selection for Human Activity Recognition}


\author{\IEEEauthorblockN{Skyler Seto*\thanks{* All authors contributed equally in this work.}}

\IEEEauthorblockA{Department of Statistical Science\\Cornell University\\
Ithaca, NY, USA\\
Email: ss3349@cornell.edu}
\and
\IEEEauthorblockN{Wenyu Zhang*}
\IEEEauthorblockA{Department of Statistical Science\\Cornell University\\
Ithaca, NY, USA\\
Email: wz258@cornell.edu}
\and
\IEEEauthorblockN{Yichen Zhou*}
\IEEEauthorblockA{Department of Statistical Science\\Cornell University\\
Ithaca, NY, USA\\
Email: yz793@cornell.edu}}

\maketitle 

\begin{abstract}
%\boldmath
Accurate and computationally efficient means for classifying human activities have been the subject of extensive research efforts.  Most current research focuses on extracting complex features to achieve high classification accuracy. We propose a template selection approach based on Dynamic Time Warping, such that complex feature extraction and domain knowledge is avoided. We demonstrate the predictive capability of the algorithm on both simulated and real smartphone data. 
\end{abstract}


% \fi
\IEEEpeerreviewmaketitle

\section{Introduction}
Wearable sensors have widespread applications in academic, industrial and medical fields. Examples of some current uses are fall detection in medical or  home settings \cite{ozdemir2014}, and Human Activity Recognition (HAR) for smart health-monitoring systems \cite{mannini2010}. 

A substantial amount of research has been done with professional on-body wearable sensors. However, these devices are usually large, impractical or inconvenient for general commercial purposes. One viable alternative is smartphone sensors. Smartphones are equipped with accelerometers and gyroscopes, which provide rich motion data on their users. Although these sensors are less accurate than professional ones, the prevalence and convenience of smartphones suggests their potential for far-reaching applications. Possible benefits include a decreased need for human supervision in medical settings, more complex human-machine interaction, and complex activity recognition on smartphone applications.

%Throughout this paper, we approach the task of smartphone motion data classification as a supervised batch learning problem with pre-labelled samples.  We use this model as proof of concept for our proposed algorithm.

Our objective is an offline implementation of smartphone motion data classification that has comparable classification accuracy and computational efficiency with current techniques, and does not require domain knowledge of HAR.

In this paper, we propose a method based on Dynamic Time Warping (DTW). DTW has recently been widely used and integrated with other methods such as decision trees \cite{rodriguez2004interval} in machine learning. Although DTW suffers from high computational costs and Dynamic Time Warping Distance (DTWD) is not a distance metric because it lacks the triangle inequality, DTW still has the potential to be a feasible answer to the above task by providing a flexible, easily interpretable time series similarity measure. By modifying DTW to improve on computational efficiency and similarity measure accuracy, we proceed to use it for motion data clustering, activity template construction and classification for our problem. Each template is the time series average representing a cluster. It has the benefit of providing visual representations of a human activity and does not require HAR knowledge for construction.  

As such, the primary contributions of this paper are:
\begin{enumerate}
\item Modification of DTW as a similarity measure for time series,
\item Procedure for template extraction in place of feature extraction.  
\end{enumerate}

Our implementation enhances the quality of prediction, avoids high dimensionality, and is robust to noise.  For demonstration, we use real human activity data, as well as synthetic data constructed from human activity data.

In Section \ref{sec: related work}, we proceed by discussing existing methods for classifying human activities using feature extraction, hierarchical divide-and-conquer strategies, and multi-modal-sequence classification. Section \ref{sec: definitions} formally defines Dynamic Time Warping and Section \ref{sec: dtwsubseq} describes our proposed modifications to increase prediction accuracy.  Section \ref{sec: tempclass} discusses our template selection and classification approach which utilizes the modified DTW.  In Sections \ref{sec: data} and \ref{sec: results}, we apply our algorithm to both real world data and synthetic data.  Finally, in Section \ref{discussion} and onwards, we compare the results of our algorithm with existing algorithms, and conclude.

\section{Related Work}
\label{sec: related work}

Considerable work has been done on classifying human activities.   We consolidate works on both professional on-body sensors and smartphone sensors, while taking note that data from the latter may need different pre-processing steps due to higher noise tendencies. Furthermore, on-body sensors can be placed at more assigned locations on the body such as arms and ankles, and may include more components such as magnetometer \cite{ozdemir2014}.

An essential step in many papers is feature extraction. Popular features for such tasks include mean, standard deviation, maximum, peak-to-peak, root-mean-square, and correlation between values of modality axes \cite{varkey2012}. Another suggested option is using autoregressive modeling to form augmented-feature vectors \cite{khan2010}. 

Following feature extraction, algorithms used for classification include Hidden Markov Models \cite{mannini2010}, Support Vector Machines \cite{varkey2012}, Multi-layer Perceptron, Naive Bayes, Bayesian Network, Decision Table, Best-First Tree, and K-star \cite{dernbach2012}. Although the strategy can increase classification accuracy, it requires manual grouping of the activities into meaningful categories. Comparative study across these algorithms has also been done \cite{altun2010comparative}.

While feature extraction is quite popular in HAR for achieving high accuracy rates, there are a few problems with feature extraction.  First, it is dependent on domain expertise as achieving high accuracy through feature extraction is only possible with the correct features. Second, feature extraction is prone to high dimensionality as a large number of features are needed.  Finally computing some of these features such as autoregression coefficients can be computationally intensive.

Current DTW methods on HAR involve aligning human behavior in video data \cite{sempena2011}. These methods use a generalized time warping mechanism to extend DTW to aligning multi-modal sequences \cite{zhou2012}. While our proposed method discussed in Section \ref{sec: dtwsubseq} uses DTW, we recognize that we cannot translate these methods directly for our approach since the data modality is different, and sequences extracted from video data may have different properties from raw sensor readings.  

\section{Definitions}
\label{sec: definitions}

In this section, we state our definition and assumptions of time series. We also introduce DTW. 

\subsection{Time Series}
\label{sec:TS}

A p-dimensional multivariate time series\\ $X_i = \{X_i(t_l) \in \mathbb{R}^p$; $l =1,\ldots,m \}$ is a sequence of data points where:  
\begin{enumerate}
\item $t_l<t_{l'}$ for $l<l'$, \\
\item $t_l \in T=[a,b]$, $\forall l=1,\ldots,m$,  \\
\item $X_i(t) \in \mathbb{R}^p, \forall t \in T$.
\end{enumerate}

We assume our training and test data are time series satisfying the assumptions above, with the additional condition that $\Delta t_i = t_i - t_{i-1} = \frac{b-a}{m-1}$. We assume that all series have the same length as observed in most data sets, but the following DTW algorithms can be easily extended to the alternate case.

In the rest of this paper, we will refer to the length of a time series as $m$ and the dimension of each point in the time series as $p$.

\subsection{Dynamic Time Warping (DTW)}
\label{sec: DTW}

DTW is an algorithm for computing the distance and alignment between two time series. It is used in applications such as speech recognition, and video activity recognition \cite{sempena2011}.

\begin{definition} \cite{muller2007}
\label{def: path}
A warping path is a sequence $w=(w_1,\dots,w_{|w|})$ where for $k \in \{1,\dots,|w|\}$, $w_k=(p_k,q_k)$ with $p_k,q_k \in \{1,\dots m\}$, satisfying the following conditions:
\begin{enumerate}
\item Boundary condition: $w_1=(1,1)$, $w_{|w|}=(m,m)$
\item Monotonicity condition: $\{p_k\}_{k=1}^{|w|}$, $\{q_k\}_{k=1}^{|w|}$ are monotonously non-decreasing sequences
\item Continuity condition: \\$w_{k+1}-w_k \in \{(1,0),(0,1),(1,1)\}$\\ for $k \in \{1,\dots,|w|-1\}$
\end{enumerate}
\end{definition}

For simplicity, we write $w=(p,q)$ as a path in accordance with Definition \ref{def: path}

\begin{definition}
\label{def: cost}
For time series $X_i$ and $X_j$ with distance matrix $D$ as calculated in Algorithm \ref{DTW}, for a path $w=(p,q)$, we define the cost $c$ as:
$$
c(w) = \sum_{k=1}^{|w|}D[p_k,q_k]
$$
\end{definition}

\begin{definition}
\label{def: dtwdef}
The Dynamic Time Warping Distance (DTWD) is the sum of the pointwise distances along the optimal path $w^*$, for the cost function defined in Definition \ref{def: cost}.  
$$
w^* = \argmin_w c(w), \quad \text{DTWD}\left(X_i,X_j\right) = c(w^*) = D[m,m].
$$
\end{definition}

DTW can be optimized through a bandwidth parameter $bw$, where it computes only values of the matrix close to the diagonal. This version of DTW is called FastDTW \cite{salvador2007toward}. Algorithm \ref{DTW} denotes the procedure to compute DTWD$\left(X_i,X_j; bw\right)$. A small bandwidth should be employed if the two time series demonstrate the same shape and frequency, since a smaller bandwidth brings the computational time closer to $O(m)$.

\begin{algorithm}
  \caption{Fast DTW}
  \label{DTW}
  %\begin{algorithmic}
  \KwData{time series $X_i, X_j$, bandwidth bw} 
  
    \KwResult{DTWD$\left(X_i,X_j;bw\right)$}
    
      Initialize distance matrix $D = 0_{m\times m}$
      $D[1, 1] = |X_i(1)-X_j(1)| $
      \For{$s \in \{2,\dots,m\}$}{ $D[s,1] = D[1,s] = \infty$}
         
         \For{$ s\in \{2,\dots,m\}$}{
         
            \For{$t \in \{\max(2, s-bw),\min(m,s+bw)\}$}{
                cost=$\min{\left(D[s-1,t],D[s,t-1],D[s-1,t-1]\right)}$
                \begin{align*}
                D[s,t] =& \left|X_i\left(s\right) - X_j\left(t\right)\right| + \mbox{cost}
                \end{align*}
                }
    }
      \textbf{return} $D[m,m]$
      
      \vspace{4mm}
   %\end{algorithmic}
\end{algorithm}

\begin{figure}[h]
\centering
\includegraphics[scale=0.5]{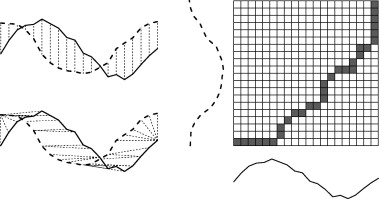}
\caption{DTW image from \cite{gorecki2015} showing the alignment procedure.  The two original time series shown on the left (dotted) and bottom (solid) of the image on the right are shown aligned on the left image according to the optimal path shown as the dark black line on the right.}
\label{fig:1}
\end{figure}

One time series $X_j$ can be aligned to another $X_i$ by using the optimal path defined in Definition \ref{def: dtwdef}. The basic concept of DTW alignment is illustrated in Figure~\ref{fig:1}. DTW first creates a matrix $D$ of pointwise distances depicted as a black and white grid in the image on the right. The algorithm then runs through $D$ from the first index (bottom left) to the last index (top right), enumerates all paths $w$, and finds an optimal warping path $w^*$ as specified in Definition \ref{def: dtwdef}. The optimal warping path is shown as the darkened line in the image on the right, and the alignment between the two series is shown in the image on the left. The algorithm also returns the sum of the pointwise distances along the optimal path \cite{gorecki2015}.

Algorithm \ref{align} is used to align two time series.  The algorithm  first computes all paths from the distance matrix D, obtains the optimal path, and updates $X_j$ according to that path.

\begin{algorithm}
  \caption{Alignment Algorithm using DTWD}\label{align}
  %\begin{algorithmic}
  \KwData{time series $X_i, X_j$, distance matrix D} 
  
    \KwResult{time series $X_j$ aligned to $X_i$}
    
    \For{$ s\in \{2,\dots,m\}$}{
         
        \For{$t \in \{\max(2, s-bw),\min(m,s+bw)\}$}{
            \If{$D[s][t-1] \leq D[s-1][t] \text{ and } D[s][t-1] \leq D[s-1][t-1]$}{
                $\text{path}[s][t] = (s, t-1)$}
            \eIf{$D[s-1][t] \leq D[s-1][t-1]$}
                {$\text{path}[s][t] = (s-1, t)$}
                {$\text{path}[s][t] = (s-1, t-1)$}

        }
    }
      $w^* = \{\}$\\
      \For{$s \in \{1,\dots m\}, t \in \{1, \dots m\}$}{
         $w^*  = w^* \cup \left\{\underset{s,t}{\argmin}\text{ path}[s][t]\right\}$
      }
        \For{$p \in w^*$}{
            $X_j\left(p[0]\right) = X_j\left(p[1]\right)$
        }
     \textbf{return} $X_j$
      
      \vspace{4mm}
      
   % \end{algorithmic}
\end{algorithm}

\iffalse
\begin{enumerate}
\item Initialize $\mathbf{D}$ as a matrix of 0s.
\item Repeat for all $i=1:m$, $j=1:n$.
\begin{enumerate}
\item Compute $cost = s_i - t_j$
\item Compute $update = \min{\left(\mathbf{D}\left[i-1,j\right],\mathbf{D}\left[i-1,j-1\right],\mathbf{D}\left[i,j-1\right]\right)}$
\item Set $\mathbf{D}\left[i,j\right] = cost + update$
\end{enumerate}
\item The DTWD is $D\left[m,n]$.
\end{enumerate}

\begin{algorithm}
  \caption{Fast Dynamic Time Warping}\label{DTW}
  %\begin{algorithmic}
  \KwData{$X_i, X_j$,window size bw} 
  
    \KwResult{DTWD$\left(X_i,X_j\right)$}
    
      \State Initialize $D = \mathbf{0_{m\times m}}$
      
      \For{$s \in \{2,\dots,m-$bw$\}$}{ \State $D[s,1] = D[1,s] = \infty$}
         
         \For{$ s\in \{1,\dots,m\}$}{
         
            \For{$t \in \{\max(0, bw-m+s),\min(m,m-t+s)\}$}{
                update$=\min{\left(D[s-1,t],D[s,t-1],D[s-1,t-1]\right)}$
                \begin{align*}
                D[s,t] =& \left|X_i\left(s\right) - X_j\left(t\right)\right| \\
                       +& \text{update}
                \end{align*}
                }
    }
      \State \textbf{return} $D[m,m]$
      
   % \end{algorithmic}
\end{algorithm}
\fi

\section{Subsequence DTW (DTWsubseq)}
\label{sec: dtwsubseq}

An apparent drawback of DTWD is the overstatement of the dissimilarity between two copies of a time series when one copy is horizontally displaced. DTW cannot match the ends of the two copies without incurring a cost. For instance, a sine curve and cosine curve can be obtained from the same sinusoidal function due to sampling from a long series, but DTWD returns a positive value. This limits the functionality of DTWD as a similarity measure.

To alleviate this, we propose a modification to DTW based on subsequence matching \cite{giorgino2009}. We relax the boundary condition that the optimal path must start at the bottom left $D[1,1]$ and end at the top right $D[m,m]$ of the distance matrix $D$. 

Computing all possible paths is computational expensive. To induce computational savings, we introduce a displacement window parameter $dw$, which is the maximum horizontal displacement of the samples. We restrict our search to paths that start and end within the window defined by $dw$. This parameter can be empirically estimated through the distribution of a common landmark in the data. For periodic data, the natural displacement window is the period.

We view similar time series as those having similar shape and frequency. Therefore, at each displacement $k$, we truncate both input series, $X_i$ and $X_j$, to the same length. To obtain a distance with respect to the original length, for the truncated series, $X^k_i$ and $X^k_j$, the resulting DTWD$\left(X^k_i,X^k_j\right)$ is weighed proportionately to their length by
$$
DTWD^k\left(X_i,X_j\right) = \frac{m}{m-k+1}DTWD\left(X^k_i,X^k_j\right)
$$

DTWsubseq, displayed below in Algorithm \ref{DTWsubseq}, is further optimized by imposing a bandwidth as in Fast DTW. 

\begin{algorithm}[h]
  \caption{Fast DTWsubseq}
  \label{DTWsubseq}
  %\begin{algorithmic}
  \KwData{time series $X_i, X_j$, displacement window dw, bandwidth bw} 
  
    \KwResult{DTWsubseqD$\left(X_i,X_j;dw,bw\right)$}
    
      Initialize $optD = \infty$
      
      \For{$k \in \{1,\dots,dw\}$}{
        $X^k_i = X_i[k:m]$ \\
        $X^k_j = X_j[1:m-k+1]$ \\
        $D_k = DTWD^k\left(X_i,X_j;bw\right)$ \\
        \If{$D_k<optD$}{$optD = D_k$}
    }
      \textbf{return} $optD$
      
      \vspace{4mm}
   % \end{algorithmic}
\end{algorithm}

\section{Template Selection and Classification}
\label{sec: tempclass}
\subsection{Overview}
As discussed in Section \ref{sec: related work}, the standard approach to HAR classification first extracts features from the motion data time series, then classifies using these features with well-established algorithms. Working with the raw data in the time domain is preferable if it is sufficiently capable of capturing information about the data.  We propose a method based on DTWsubseq and hierarchical clustering, which avoids intensive feature extraction, and can alleviate many of the problems with feature extraction discussed in Section \ref{sec: related work}.

Our approach to classification is to use the training data to build time series templates representing each activity, and subsequently classify the test data according to their similarity to these templates using DTWsubseqD. This type of template-based method is robust to speed and style variations of the subjects' motions, and potentially requires less training data than feature-based methods  \cite{sempena2011}.

\begin{figure}[h]
\centering
\includegraphics[scale=0.25]{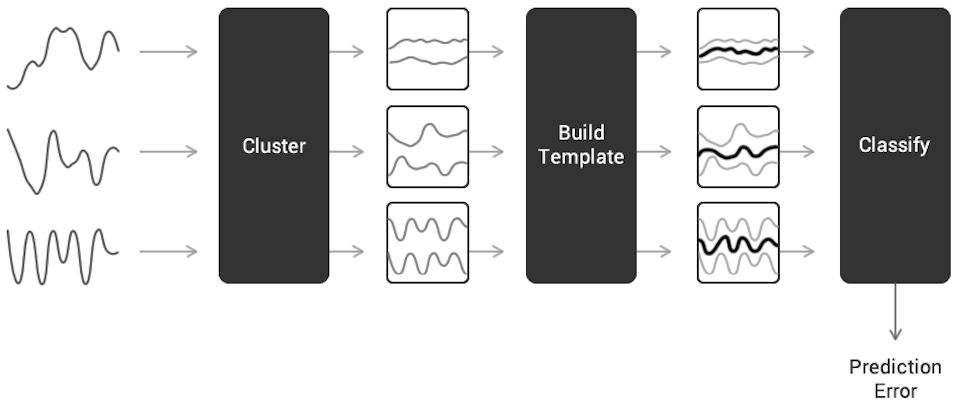}
\caption{Breakdown of the template selection and classification method}
\label{fig:workflow}
\end{figure}

Before giving details of how to construct templates, we present a general diagram of the full classification procedure illustrated in Figure \ref{fig:workflow}.

\subsection{Cluster}

For each activity, we use the training set to build clusters.  There are two main parts of the clustering step:
\begin{enumerate}
\item Computing the distance matrix by finding the DTWsubseqD between all pairs of points in each activity,
\item Forming clusters in hierarchical clustering by calculating distance between two clusters $C_i$ and $C_j$ is calculated as $$d(C_i,C_j) = \max_{s\in C_i,t\in C_j} DTWsubseqD(s,t),$$ and removing flat clusters by restricting pairwise distances within a cluster to be below $$cut \times  \max_{C_i,C_j}d(C_i,C_j).$$

\end{enumerate}

We use hierarchical clustering as our clustering algorithm because of its flexibility in being able to perform with any similarity measure. Decreasing the parameter $cut$ increases the number of clusters for each activity.  We show results for different values of $cut$ in Section \ref{sec: results}.

\subsection{Build Templates}

After computing clusters for each activity, we compute the average of a set of time series.  We consider two methods for selecting templates: DTW Pointwise Averaging (DPA) and DTW Barycenter Averaging (DBA).

We provide the algorithm for DPA in Algorithm \ref{DPA} and brief descriptions of DBA. 

DPA is a straightforward method for computing the average of a set of time series. The algorithm first finds the point with minimum distance to all other time series, aligns each time series to that point, and finally calculates a pointwise average.  

\begin{algorithm}
  \caption{DTW Pointwise Averaging}\label{DPA}
  %\begin{algorithmic}
  \KwData{Clusters $C_1, \dots, C_n$} 
  
    \KwResult{Templates $T_1,\dots, T_n$}

      \For{$i \in \{1,\dots,n\}$}{
        $t^* = \underset{t}{\min}\underset{x \in C_i}{\sum}DTWsubseqD\left(x,t\right)$\\
        \For{$x \in C_i$}{
            $\text{align}(t^*,x)$\\
            
            $T_i = \|C_i\|^{-1} \underset{x \in C_i}{\sum}x$
        }
    }
     \textbf{return} $T_1,\dots, T_n$
\vspace{4mm}      
   % \end{algorithmic}
\end{algorithm}

\iffalse
\begin{enumerate}
\item Find $t^*$, the time series in the cluster with minimum distance to all other time series.
\item Repeat for each cluster.
\begin{enumerate}
\item Align* the remaining time series in the cluster to $t^*$.
\item Compute the pointwise average of all time series.
\end{enumerate}
\item The averages for all clusters are the templates.
\end{enumerate}
\fi

DBA is a global method that calculates an average series that minimizes the sum of squared DTW distances to all series in the cluster. In each iteration of the algorithm, 
\begin{enumerate}
\item Conduct DTW between the average series and each series in the cluster, and extract associations between the coordinates of the pair.
\item For each coordinate $\alpha$ of the average series, update the corresponding value as the mean corresponding to all coordinates associated with $\alpha$.
\end{enumerate}

Details of the DBA algorithm can be found in \cite{petitjean2011}. DBA is more computationally intensive than DPA, but DBA has been shown to improve classification accuracy in centroid-based methods \cite{petitjean2014}, and hence is included for comparison.

In the aligning and averaging portions of both algorithms, we used DTW instead of DTWsubseq because DTW would return similar warping paths while maintaining the length of the input series.

\subsection{Classify}
\label{classify}
For each training and test sample, we first compute DTWsubseqD to each template.  This provides a vector of distances, which is the same size as the number of templates, for each sample.  We treat this vector of distances as a feature vector, and proceed by running a dimensionality reduction algorithm (such as PCA) and a classifier (such as SVM) on the DTWsubseqD vectors.

\section{Real World Data and Simulations}
\label{sec: data}

\subsection{UCI HAR Data}
The dataset contains a total of 10,299 samples from 30 subjects conducting six activities in a lab: 3 dynamic activities (walking, walking upstairs, walking downstairs) and 3 static activities (sitting, standing, lying). We used 7,352 samples for training, and the remaining 2,947 samples for testing.  

Each reading (time series) is six dimensional with dimensions corresponding to acceleration (in ms${^{-2}}$), and angular velocity (in s$^{-1}$) in the x-y-z axis.  The range for the acceleration is $[-1,1]$ and the range for angular velocity is $[-3,3]$.

\begin{figure}
\centering
\begin{tabular}{c c}
\includegraphics[scale=0.2]{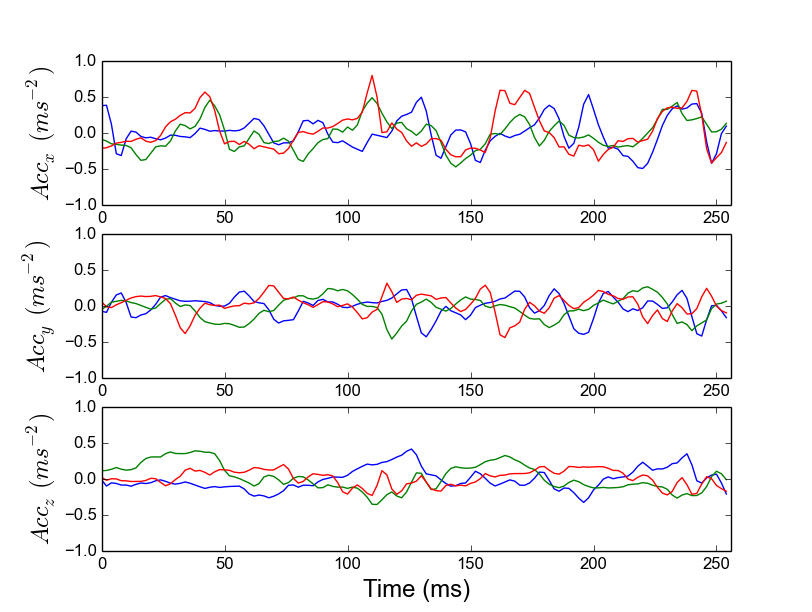} &  \includegraphics[scale=0.2]{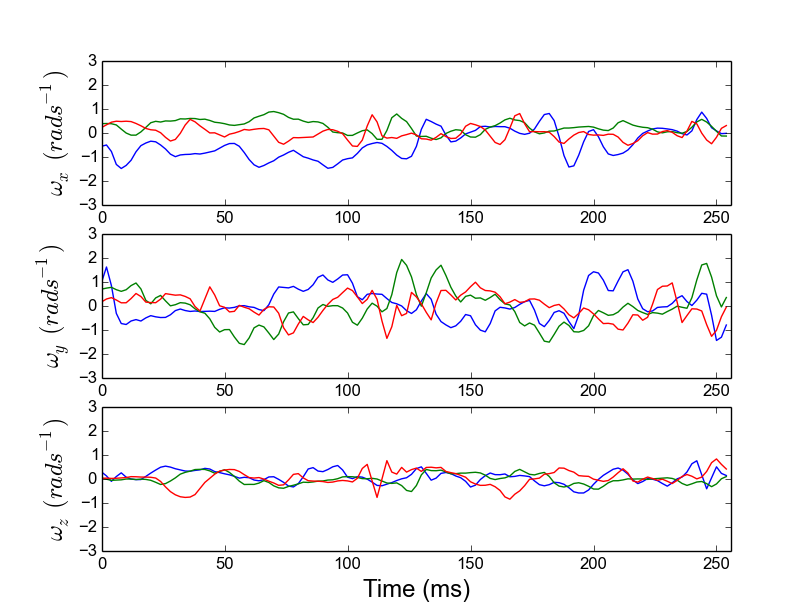}\\
{\scriptsize Dynamic Activities Acceleration} & {\scriptsize Dynamic Activities Angular Velocity } \\
\includegraphics[scale=0.2]{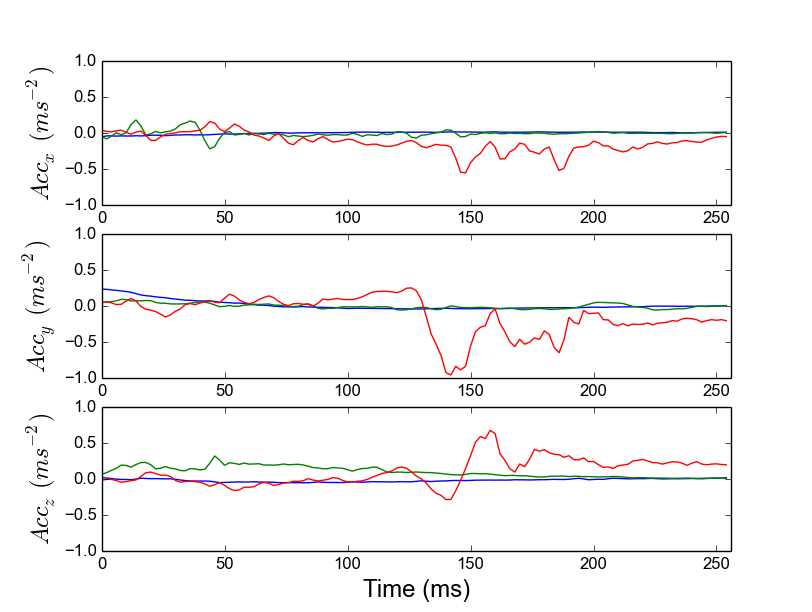} & \includegraphics[scale=0.2]{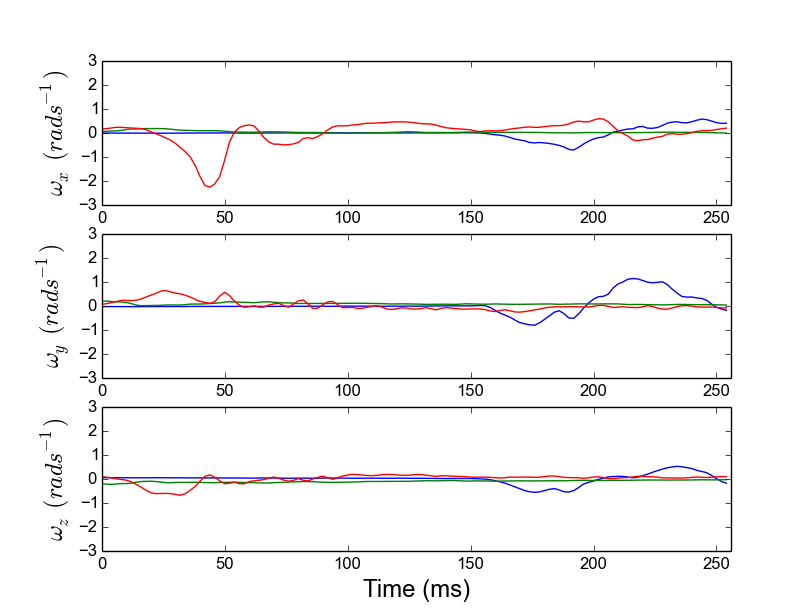} \\
{\scriptsize Static Activities Acceleration } & {\scriptsize Static Activities Angular Velocity } 
 
\end{tabular}
\caption{A sample of pre-processed curves from the UCI HAR dataset.  Each figure contains 3 plots corresponding to the x,y, and z axis (top to bottom). The top pair of plot has 3 curves corresponding to the dynamic activities: walking (blue), walking upstairs (green), walking downstairs (red), and the bottom pair of plots has 3 curves corresponding to the static activities: sitting (blue), standing (green), lying (red).}
\label{fig:curves}
\end{figure}

This data set is already preprocessed as provided by applying noise filters and then sampled in fixed-width sliding windows of 2.56 sec and 50\% overlap (128 readings/window) \cite{anguita2013}. Figure \ref {fig:curves} depicts some curves after preprocessing.

\begin{figure}
\centering
\begin{tabular}{c c}
\includegraphics[scale=0.2]{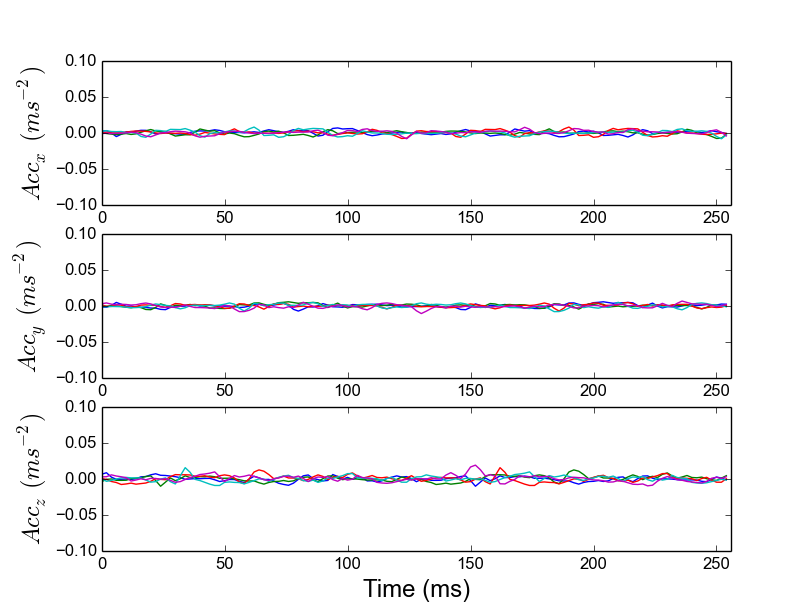} &  \includegraphics[scale=0.2]{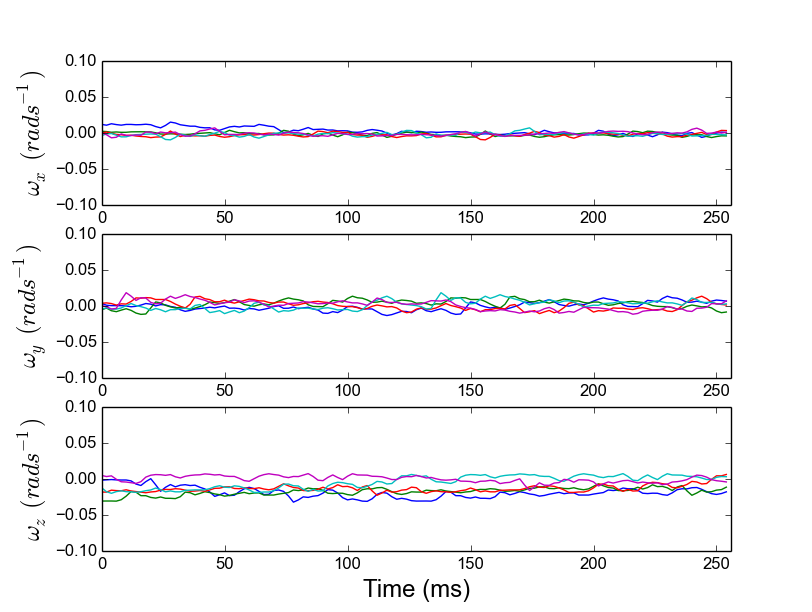}\\
{\scriptsize Activity 3 Acceleration} & {\scriptsize Activity 3 Angular Velocity } \\

\iffalse
\includegraphics[scale=0.2]{xTrainF_acc4.png} & \includegraphics[scale=0.2]{xTrainF_gyr4.png} \\
{\scriptsize Activity 4 Acceleration } & {\scriptsize Activity 4 Angular Velocity } \\
\includegraphics[scale=0.2]{xTrainF_acc5.png} & \includegraphics[scale=0.2]{xTrainF_gyr5.png} \\
{\scriptsize Activity 5 Acceleration } & {\scriptsize Activity 5 Angular Velocity } 
\fi
\end{tabular}
\caption{A sample of flat curves removed from the UCI dataset from the sitting activity.  Each figure contains 3 plots corresponding to the x,y, and z axis (top to bottom). }
\label{fig:flat}
\end{figure}

Due to the method of sampling, many of the resulting shortened curves in the static activity category are flat curves with no activity, and mis-informative for classification. Since these flat curves provide no additional information, we have removed them.  A multidimensional time series will be considered as a flat curve if all of its dimensions are flat.  A dimension of the time series is flat if its range (maximum value - minimal value) falls within the 5\% quantile of all ranges of that dimension across all time series.  Figure \ref{fig:flat} depicts some curves which have been removed. 

\subsection{Synthetic Data}

This dataset is created to test the ability of our proposed method to classify time series belonging to new subjects, in the event of noise that may be present in this type of usage of smartphone sensors. In the simulation, we consider short bursts of noise in the observations, due to reasons such as jerky motions, sudden shifts of the sensor, and sensor noise. 

Training data is generated from one dimension ($Acc_x$, Acceleration in the x-axis) of a random template, computed from the UCI HAR dataset, from each of the following activities: walking, walking upstairs, walking downstairs, sitting. This can be taken to represent the motion of one subject. Test data is generated from a different random template for each activity, to simulate motions from a different subject. 

The training set has a total of 800 samples, 200 for each activity. The test set has 200 samples,  50 for each activity. Each sample is created from the template in the following manner:
\begin{enumerate}
\item $Acc_x$ of the template is extracted and normalized to zero mean and unit variance
\item Concatenate the template to length 256
\item Perform FFT on the resulting series 
\item Generate a normally distributed $\mathcal{N}(0,5)$ random vector of length 10 and add it to a random location of the FFT
\item Perform IFFT and retrieve the real part of the result
\item Sample a series of length 128
\end{enumerate}

\begin{figure}
\centering
\begin{tabular}{c c}
\includegraphics[scale=0.2]{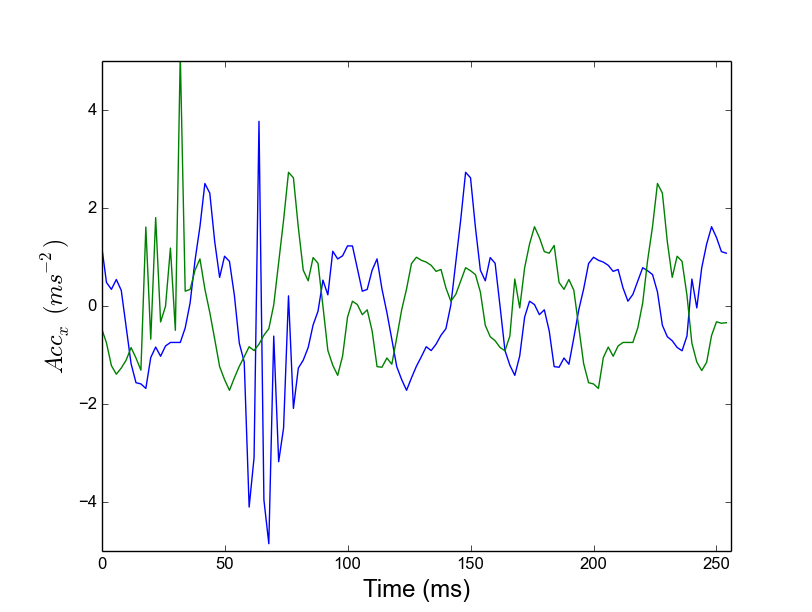} &  \includegraphics[scale=0.2]{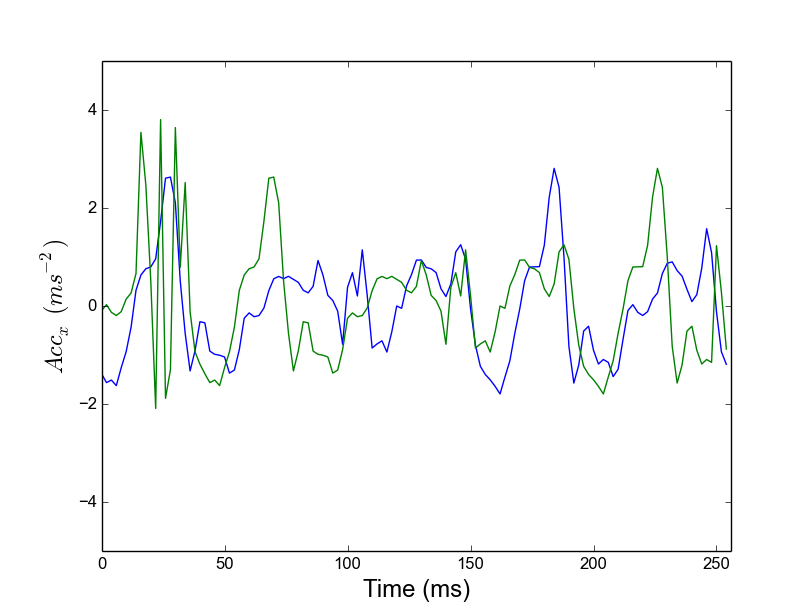}\\
{\scriptsize Training Sample for Walking} & {\scriptsize Test Sample for Walking}
\end{tabular}
\caption{Training and Test Samples from the Synthetic Dataset for Walking}
\label{fig:syn2}
\end{figure}

Figure \ref{fig:syn2} contains two plots corresponding to training and test samples for the same activity.  Two curves are plotted to show the effects of adding noise.  Each reading (time series) is a one dimensional acceleration (in ms${^{-2}}$).  The range for the acceleration is $[-4,4]$.

\section{Results}
\label{sec: results}
In this section, we present the results on the UCI HAR dataset and synthetic data.

\begin{figure}
\centering
\begin{tabular}{c c}
\includegraphics[scale=0.2]{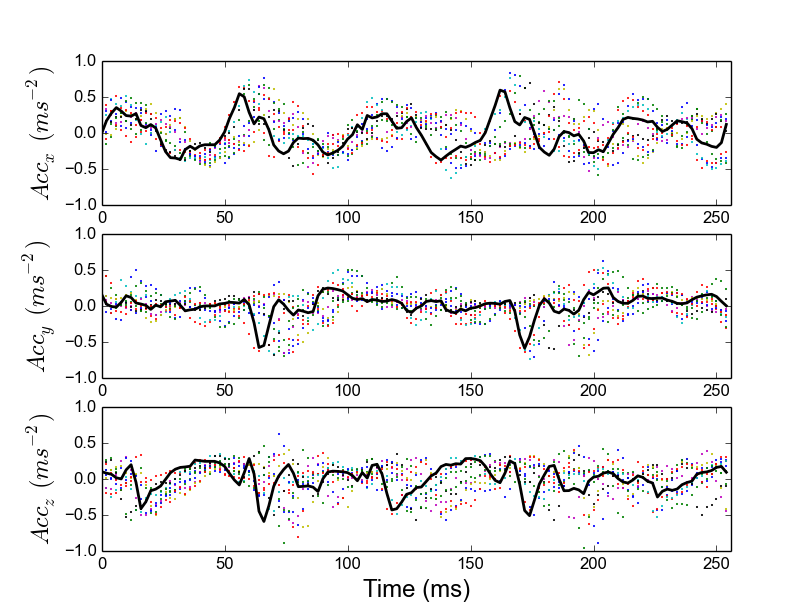} &  \includegraphics[scale=0.2]{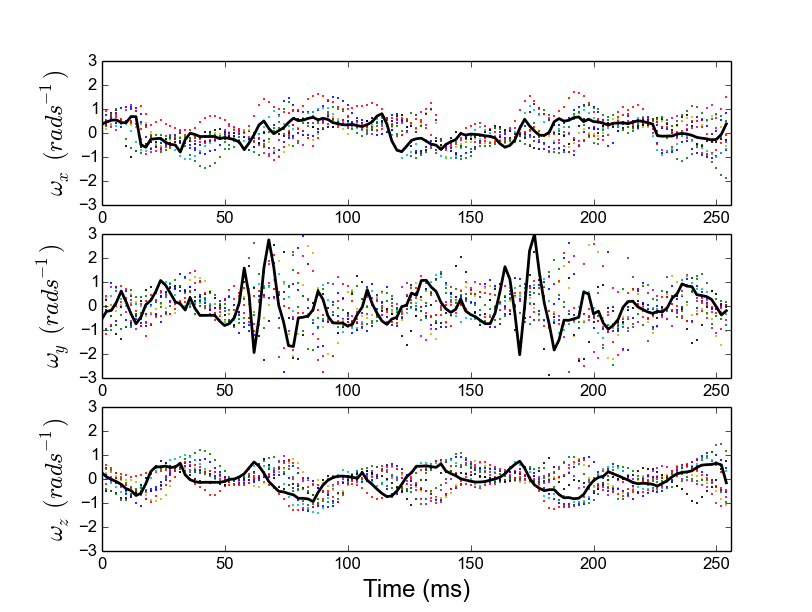}\\

{\scriptsize DBA Cluster 1  Acceleration} & {\scriptsize DBA Cluster 1 Angular Velocity } \\

\includegraphics[scale=0.2]{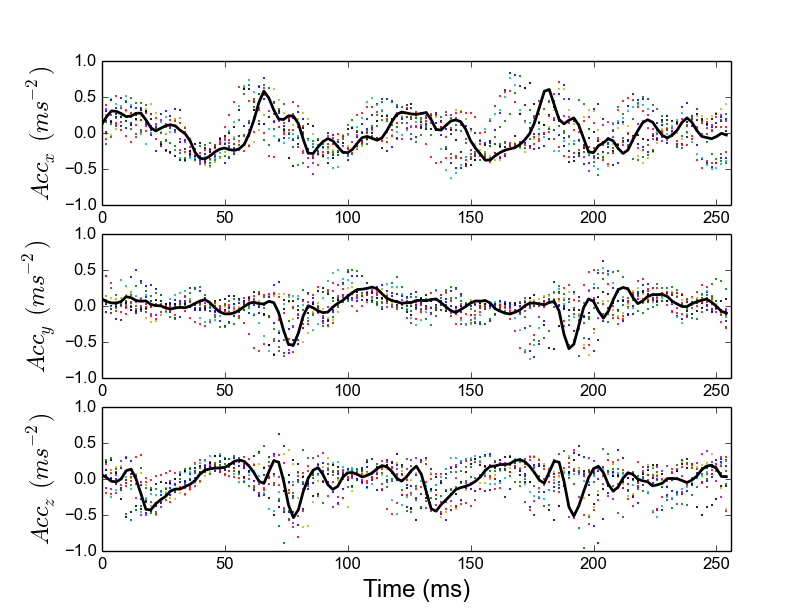} & \includegraphics[scale=0.2]{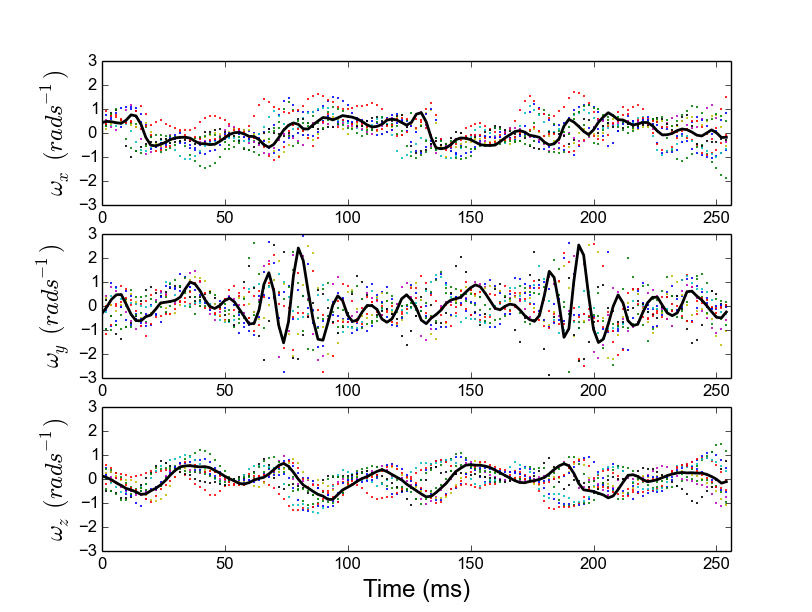} \\

{\scriptsize DPA Cluster 1  Acceleration } & {\scriptsize DPA Cluster 1 Angular Velocity } \\

\includegraphics[scale=0.2]{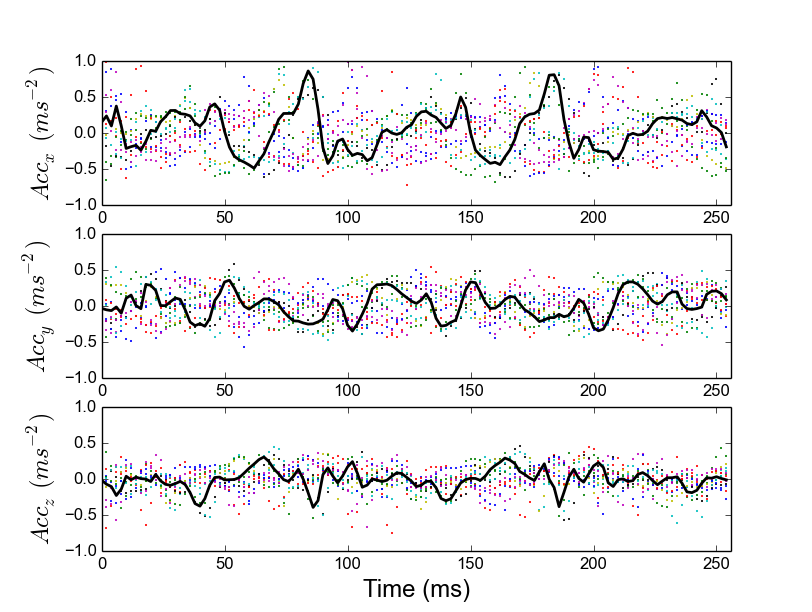} & \includegraphics[scale=0.2]{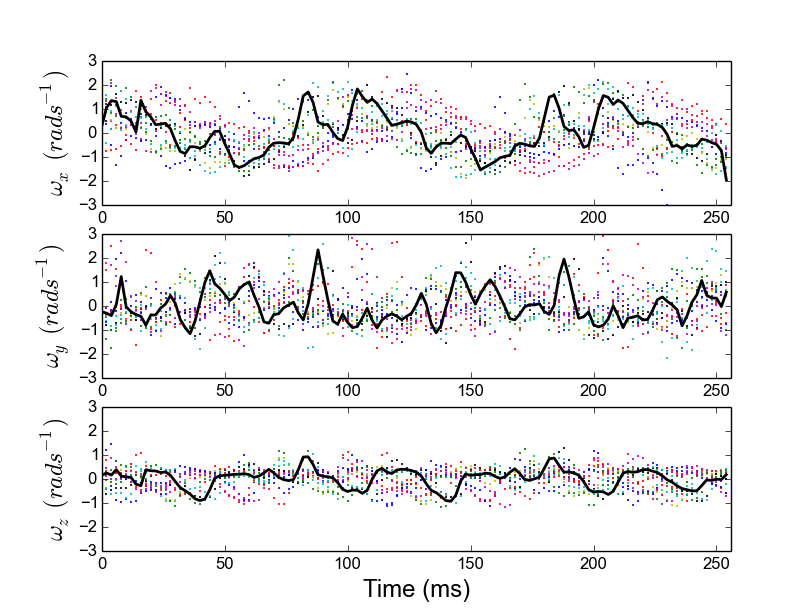} \\

{\scriptsize DBA Cluster 2 Acceleration } & {\scriptsize DBA Cluster 2 Angular Velocity } \\

\includegraphics[scale=0.2]{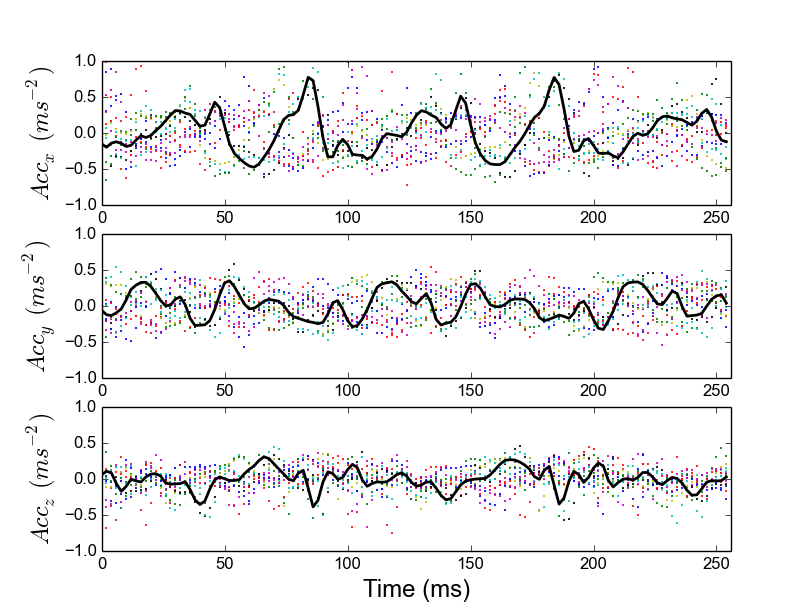} & \includegraphics[scale=0.2]{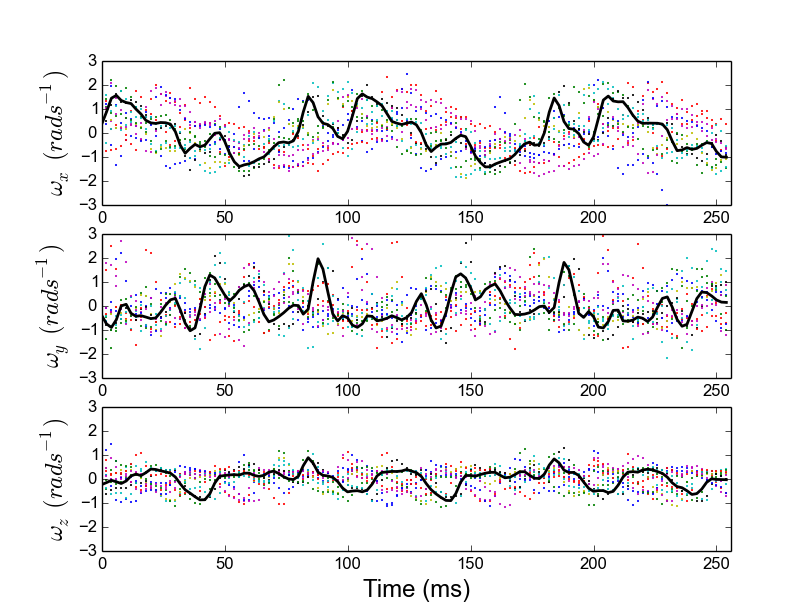} \\

{\scriptsize DPA Cluster 2 Acceleration } & {\scriptsize DPA Cluster 2 Angular Velocity } 
\end{tabular}
\caption{UCI HAR Data: Plots for templates constructed from two clusters for walking.  Both DBA and DPA are shown for comparison.  Each figure contains 3 plots corresponding to the x,y, and z axis (top to bottom).}
\label{fig:templates}
\end{figure}

\begin{figure}
\centering
\begin{tabular}{c c}
\includegraphics[scale=0.20]{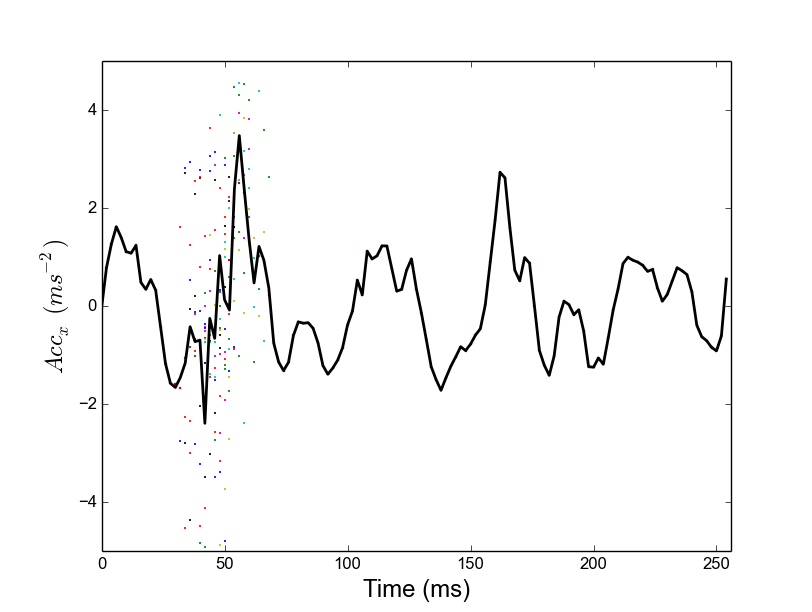} &  \includegraphics[scale=0.20]{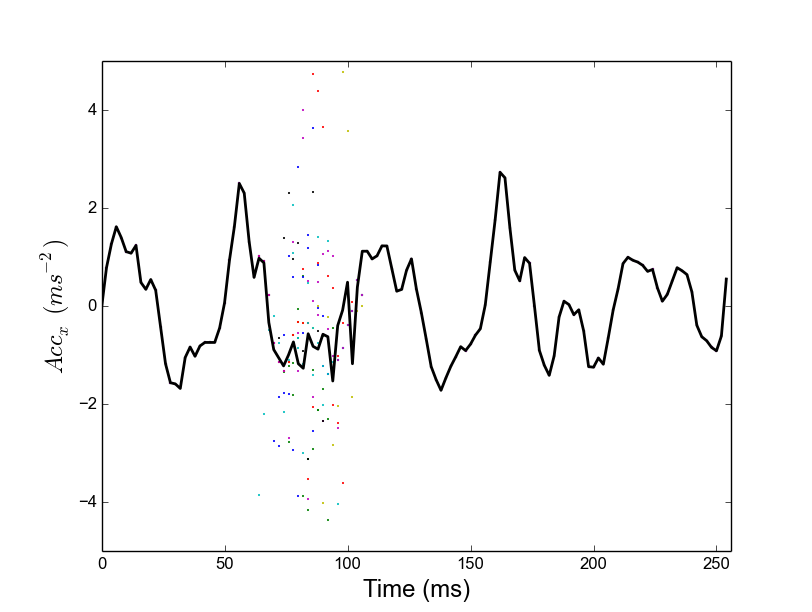}\\

{\scriptsize DBA Cluster 1  } & {\scriptsize DBA Cluster 2 } \\

\includegraphics[scale=0.20]{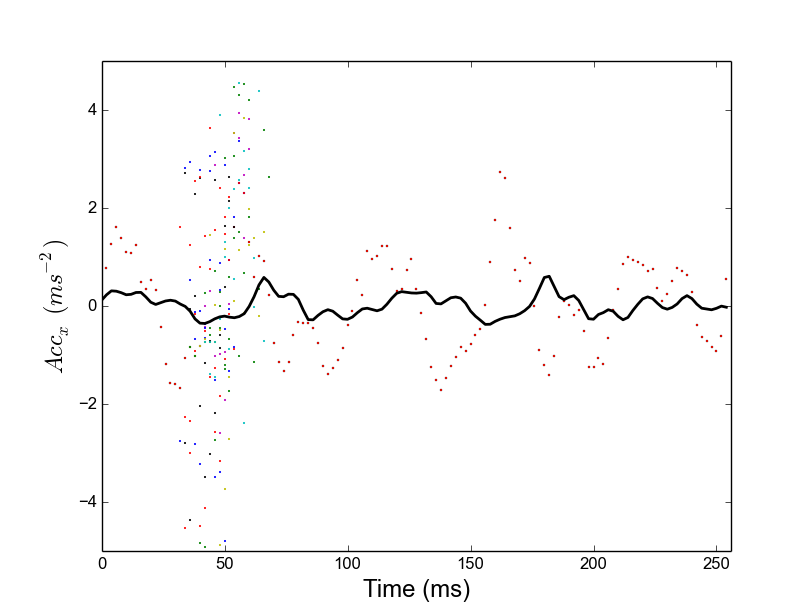} & \includegraphics[scale=0.20]{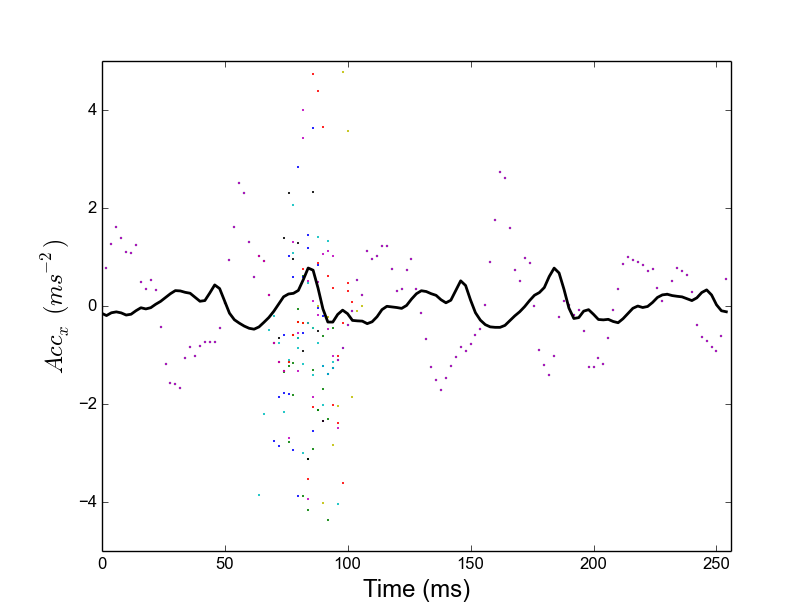} \\

{\scriptsize DPA Cluster 1   } & {\scriptsize DPA Cluster 2} 

\end{tabular}
\caption{Synthetic Data: Plots for templates constructed from two clusters for walking $Acc_x$.  Both DBA and DPA are shown for comparison.}
\label{fig:templatesyn}
\end{figure}

We examine templates generated by our method, and additionally study the benefit obtained from using different values of the $cut$ parameter in the clustering stage.  We plot templates corresponding to different clusters in the same activity to demonstrate the performance of our clustering and template construction methods.  Templates are shown in Figure \ref{fig:templates} and \ref{fig:templatesyn} for both DBA and DPA for the UCI dataset, and synthetic dataset.

Classification accuracy is presented in Tables
\ref{lab:accuracyUCI} 
and \ref{lab:accuracySyn}.  Each table shows the accuracy for all four combinations of distance method (DTW, DTWsubseq) and averaging method (DPA, DBA), and two different values of the $cut$ parameter (0.25, 0.5). 

 \begin{table}[h!]
  \centering

 \begin{tabular}{c|c|c}
   cut & DTWsubseq with DPA & DTWsubseq with DBA\\
   \hline
   0.5 & 0.789 (0.946) &  0.777 (0.941)\\
   0.25 & 0.838 (0.974) & 0.855 (0.977) \\

\multicolumn{3}{c}{\vspace{5mm}}\\

  cut & DTW with DPA  & DTW with DBA \\
  \hline
  0.5 & 0.781 (0.943) &  0.797 (0.937)  \\
  0.25 & 0.841 (0.966)  & 0.860 (0.977) 
  \end{tabular}
  
  \vspace{4mm}
  
  \caption{Test Accuracy for  DTW and DTWsubseq UCI HAR Dataset.  Numbers in parentheses are the accuracies after combining all static activities into one activity.}
  \label{lab:accuracyUCI}

\end{table}

 \begin{table} [h!]
  \centering
  
  \begin{tabular}{c|c|c}
   cut& DTWsubseq with DPA & DTWsubseq with DBA \\
   \hline
   0.5&   0.650 &  0.640  \\
   0.25& 0.700 & 0.655 \\

\multicolumn{3}{c}{\vspace{5mm}}\\

  cut \hspace{-4mm}&\hspace{6mm} DTW with DPA \hspace{0mm} & DTW with DBA \\
  \hline
  0.5 & 0.615  &  0.620 \\
  0.25 & 0.615 & 0.580 
 
  \end{tabular}
  
  \vspace{4mm}
  
  \caption{Test Accuracy for  DTW and DTWsubseq Synthetic Dataset.}
  \label{lab:accuracySyn}
\end{table}

\section{Discussion}
\label{discussion}

\subsection{Comparison with Feature Extraction}
We benchmark our procedure against standard feature extraction methods.  We computed 568 features using mean, standard deviation, correlation, RMSE from \cite{varkey2012}, Energy \cite{ravi2005}, average absolute difference \cite{kwapisz2011activity}, largest 5 FFT magnitudes, autocorrelation, kurtosis, skew from \cite{altun2010comparative},  autoregression coefficients \cite{khan2010}, and number of zeros for the original time series, and first difference time series.  We also computed the mean, standard deviation, kurtosis, and skew for the magnitudes of the FFT \cite{anguita2013}.  

We performed standard classification on these features using PCA for dimensionality reduction and linear SVM as a classifier (same as in Section \ref{classify}).  The results for classification on both synthetic data and the UCI dataset are shown in Table \ref{tab:accf}.

 \begin{table}[h]
  \centering
  \begin{tabular}{c | c }
  UCI Feature & Synthetic Feature \\
  \hline \\
  0.890 (0.965) & 0.67

  \end{tabular}
  \vspace{4mm}
 \caption{Accuracy using Feature Extraction.  Number in parentheses is the accuracy after combining all static activities into one activity.}
  \label{tab:accf}
 \end{table}

Before comparing accuracy, we comment on a few advantages our method has over feature extraction.  First, our methods have the natural advantage that no domain knowledge is required, compared to the importance of extracting the correct features.  In order to achieve high accuracy using feature extraction, a large number of features need to be extracted (over 500).  To achieve comparable accuracy, using DTW-DBA with cut $= 0.25$ our method builds around $220$ templates, less than half the number of features.  Second, some features such as autocorrelation increase with the length of the time series, and almost all features increase with $p$, the dimension of each point.  The dimensionality of our method is only influenced by the number of templates, which can be fixed through the $cut$ parameter.  Third, our method has reduced computational complexity compared to feature extraction.  Fast DTW has complexity $O(m)$, whereas most features have computational complexity at least $O(m)$, and some features such as autoregression coefficients have computational complexity $\Omega\left(m^2\right)$.

\subsection{UCI HAR Data}

Figure \ref{fig:templates} shows that our clustering scheme performs reasonably well in grouping differently-shaped curves. Both template methods are able to capture the shape of the data. In fact, the DPA and DBA templates appear very similar to one another. 

Judging from the classification accuracy results in Table \ref{lab:accuracyUCI}, for both $cut = 0.5$ and $0.25$, DTW performs better with DBA, achieving 0.797 over 0.781 and 0.860 over 0.841 respectively. When $cut = 0.25$, DTWsubseq also does better with DBA, with accuracy 0.855 over 0.838. But when $cut = 0.5$, DTWsubseq performs better with DPA, achieving 0.789 over 0.777. 

One possible reason why DBA did not consistently work better is that the starting average sequence in DPA is initialized using the same similarity measure (DTWD, DTWsubseqD) as is used for clustering. The starting average sequence in DBA is initialized with a random candidate in the cluster, as suggested in the original paper \cite{petitjean2011}.

It's possible that with a few modifications, such as picking the right initializing series for DBA, DTWsubseq-DBA with $cut = 0.5$ could achieve higher accuracy as it performs well in distinguishing between static and dynamic activities. It makes only 22 classification errors between static and dynamic activities, comparable to DTWsubseq-DPA, and has comparable accuracy when considering 4 activities as shown in parentheses in Table \ref{lab:accuracyUCI}.

The effects of lowering the $cut$ threshold are clear-cut.  As expected, lowering the threshold (and increasing the number of clusters) increased accuracy .  With original DTW we see an increase in accuracy from 0.781 to 0.841 with DPA, and subsequently up to 0.86 with DBA.  Using DTWsubseq, we see an increase with DBA, and a smaller increase with DPA.  This may primarily be due to overfitting.  Decreasing cut from 0.5 to 0.25 increase the number of templates from around 50 to over 500.  

The highest accuracy our method achieved was 0.860 using DTW-DBA with $cut = 0.25$, which is comparable to the feature extraction method which had a test accuracy of 0.890.  One possibility is that the flat curves in static activities have not been sufficiently removed. The static activity curves usually have a spike, then remain flat. As some flat portions still remain in our training and test sets, all 3 static activities contain flat templates and hence induce high classification error within themselves. This is evidenced by the confusion matrix for DTW-DBA shown in Table \ref{cm} where there are lots of mis-classifications between activities 3-5, especially for activity 5.  

\begin{table}[h]
{\small \begin{center}
\hspace{17mm} Predicted 
$$\text{Actual } \kbordermatrix{
    & 0 & 1 & 2 & 3 & 4 & 5 \\
 0 & 462 &  26 &  8 &   0 &   0 & 0\\
  1 & 3& 460 &  8 &   0 &   0 &   0\\
  2 & 0 &    1 &  419 &    0 &    0 &   0\\
  3 &  0&   1&   0& 182&  32&   5\\
  4 & 2&   0&   1&  71& 258&   5\\
  5 & 0&   0 &0 &  75&  63&  71
  }
  $$
  \end{center}
  }
  \caption{Classification by DTWsubseq-DBA and $\text{cut}= 0.25$.  Activities are labeled as follows: 0- walking, 1 - walking upstairs, 2 - walking downstairs, 3 - sitting, 4 - standing, 5 - lying}
  \label{cm}
  \end{table}

However, we find that our methods are capable of performing better than feature extraction when we circumvent the issue of the flat templates. Looking at the values in parenthesis in Table \ref{lab:accuracyUCI} where we perform a 4-category classification of walking, walking upstairs, walking downstairs and static activities, all our methods with $cut = 0.25$ outperforms feature extraction's 0.965 accuracy. DTW-DBA and DTWsubseq-DBA perform the best with 0.977.

Summarizing the results of our method on the UCI HAR dataset, compared with benchmark feature extraction method, we see that our accuracy rates are  similar for six activities and better for four activities.  Within our template-based methods, we see that there is no clear improvement using DPA versus DBA, and there is a slight overall improvement using DTWsubseq over DTW.  Finally, we see that in all cases decreasing the $cut$ parameter increases accuracy.

\subsection{Synthetic Data}

Our synthetic data experiments reflect the satisfactory performance of our methods in the event of noisy data and new test subjects not present in the training phase. 

Comparing template construction, Figure \ref{fig:templatesyn} shows that templates created using DBA are significantly more robust to sporadic noise than those created by DPA. The samples in  Figure \ref{fig:templatesyn} are created from a single time series and modified by adding noise at a  small region of the curve.  DBA successfully accounts for this, as its template is identical to all curves at points other than the noisy region, and its shape matches the noisy region fairly well.  On the other hand, DPA fails to capture the identical shape of the curve as it underestimates the magnitudes of the peaks and troughs.

However, because the test data is constructed from a different template (simulating a different person), DBA overfits to the training data. All methods with DPA perform similarly or better than with DBA. With DPA instead of DBA, DTWsubseq achieves 0.650 over 0.640 with $cut=0.5$, and 0.700 over 0.655 with $cut=0.25$. 
DTW performs similarly with DPA and DBA with 0.615 and 0.620 respectively, and better with DPA for $cut=0.25$ obtaining 0.615 over 0.580.  

The $cut$ threshold follows similar trends as in the UCI dataset, except that $cut=0.25$ for DTW-DBA achieves lower accuracy than $cut=0.5$.  This is likely due to overfitting as the number of templates created was around 400.

Our method achieves the highest accuracy using DTWsubseq-DPA with $cut = 0.25$ (0.700),  3\% higher than that of feature extraction (0.67).  This gives good indication that our method is potentially suitable for real-world applications, since it is more robust to noise than feature extraction, and is better at generalizing to data from new test subjects. 

Summarizing the results of our method on the synthetic dataset, we see our method achieves higher accuracy than benchmark feature extraction. Within the template-based methods, we again see that there is no clear improvement using DPA versus DBA, but there is an  improvement using DTWsubseq over DTW.  Finally, we see that in almost all cases, decreasing the $cut$ parameter increases accuracy as expected.

\subsection{Summary of Results}
We summarize the accuracy results obtained using our algorithm for clarity in Table \ref{lab:summary}.  We see that our algorithm's accuracy is comparable to feature extraction's, and performs better in most cases.

 \begin{table} [h!]
  \centering
  
  \begin{tabular}{c|c|c}
    &Accuracy on six activities & Accuracy on four activities \\
   \hline
   UCI HAR    &  0.860 (0.890) & 0.977 (0.965) \\
   Synthetic & N/A & 0.700 (0.67) 
 
  \end{tabular}
  
  \vspace{4mm}
  
  \caption{Summary Table for Best template selection algorithm vs. feature extraction.  Feature extraction accuracy is shown in parentheses.}
  \label{lab:summary}
\end{table}

\section{Conclusion}
\label{sec: conclusion}
In this paper, we presented modifications to DTW as a more accurate time series similarity measure, as well as a template-based approach for Human Activity Recognition. Through experiments on both real data and synthetic data, we show that our approach gives comparable and sometimes even better classification accuracy than the most common method of feature extraction. As compared to feature extraction on real data, our approach has comparable overall test accuracy, and better accuracy within activities in the dynamic category. For the synthetic data, it achieved higher accuracy, indicating robustness to noise and the ability to classify data from new test subjects. This is an especially useful feature for real-life implementations of such classification algorithms, for instance on a smartphone app.

Furthermore, the application of our approach extends beyond HAR, as long as the data set satisfies our definition of multi-dimensional time series. Since the template-based approach allows us to extract features without domain knowledge, it can be readily applied to these new datasets, unlike feature extraction methods which requires careful determination of useful features to achieve good accuracy.

\iffalse
Time series-based classification and feature-based methods, each with their benefits and drawbacks, have been used to classify human activities. In this paper, we proposed and evaluated a multi-stage process for classifying human activities by directly comparing the time series.. Using synthetic examples and a real data set, we demonstrate the applicability of our approach in its ability to achieve high accuracy rates close to 90\%, its potential to achieve higher computational efficiency.

\fi
\section{Future Work}
\label{sec: future}
We plan to improve our activity recognition and template usage in several ways.  Potential directions include: modifying DTW to learn more complex activities, cross validating parameters in hierarchical clustering and DTW such as $cut$ or bandwidth $bw$ to increase prediction accuracy, removing redundant templates and limiting overfitting, and creating synthetic data from templates which can increase predictive power, or be used in place of real-world data.

Since the work presented in this paper is not specific to HAR applications,  we also plan to apply and evaluate our algorithm on sequential data in other domains of research.




%\bibliographystyle{IEEEtran}
\iffalse

\fi

%\bibliographystyle{IEEEtran}
\bibliography{main}

\end{document}